**Title:** Causal Machine Learning Is Not a Panacea: A Roadmap for Observational Causal Inference in Health

**Authors and affiliations:** Donna Tjandra (1)*, Trenton Chang (1)*, Sonali Parbhoo (2), Rajesh Ranganath (3 and 4), Andre Kurepa Waschka (5), William Mitchell (6), Maggie Makar (1), Shalmali Joshi (7), Finale Doshi-Velez (8), Leo Anthony Celi (9, 10, and 11), Jenna Wiens (1)
((1) Division of Computer Science and Engineering, University of Michigan, Ann Arbor, Michigan, United States,
(2) Department of Electrical and Electronic Engineering, Imperial College London, London, UK,
(3) Courant Institute of Mathematical Sciences, New York University, New York, New York, United States,
(4) Center for Data Science, New York University, New York, New York, United States,
(5) Department of Mathematics & Statistics, Elon University, Elon, North Carolina, United States,
(6) Department of Ophthalmology, Cambridge University Hospitals, Cambridge, UK,
(7) Department of Biomedical Informatics, Columbia University, New York, New York, United States,
(8) School of Engineering and Applied Science, Harvard University, Cambridge, Massachusetts, United States,
(9) Laboratory for Computational Physiology, Institute for Medical Engineering and Science, Massachusetts Institute of Technology, Cambridge, Massachusetts, United States,
(10) Department of Medicine, Beth Israel Deaconess Medical Center, Boston, Massachusetts, United States,
(11) Department of Biostatistics, Harvard T.H. Chan School of Public Health, Boston, Massachusetts, United States)
*Equal contribution

**Corresponding Author:**Jenna Wiens, PhD, Division of Computer Science and Engineering, University of Michigan, 2260 Hayward Street, Ann Arbor, MI, 48109 (Email: wiensj@umich.edu, telephone: 734-647-4832)

## Abstract (150/150 words)

**Objective:** The growing availability of large-scale observational clinical datasets and challenges in conducting randomized controlled trials have spurred enthusiasm in using causal machine learning (ML) for causal inference in observational data. We present a roadmap for applying causal ML to observational data.

**Materials and methods:** We outline the importance of assessing validity assumptions within available data and applying causal ML responsibly for clinical experts using causal ML and ML practitioners with limited clinical expertise.

**Observations:** Despite advances in causal ML, its limitations remain largely under-appreciated across disciplines. This gap in shared knowledge may impact the validity of findings.

**Discussion:** Causal assumptions must be satisfied and modeling choices justified. Otherwise, these approaches risk producing biased or misleading results, with consequences for clinical research and patient care.

**Conclusion:** Causal ML can be a powerful tool for generating causal hypotheses. We provide a template to strengthen the rigor and interpretability of causal analyses.

## INTRODUCTION

As large-scale, multi-modal observational health data become increasingly available, advances in machine learning (ML) present opportunities to generate novel clinical insights.[1-8] The combination of these rich data and powerful modeling approaches improves our ability to pursue a central goal of clinical research: identifying causal relationships. Causal inference is a tool for answering "what if" questions: *what is the outcome (eg, recovery from disease) if we administer the treatment (eg, medication)?* Causal inference can inform general guidelines and quantify population-level trends.[9] For example, retrospective analyses on historical data may help answer: *did the treatment lead to increased patient recovery in some population?* Causal inference can also inform personalized insights. For example, developing a model to predict conditional average treatment effects may help answer: *would treating a patient with these characteristics work?* Regardless of the use case, the field of causal ML adapts standard ML techniques for causal inference.

While randomized controlled trials (RCTs) remain the gold standard for establishing causal relationships, they are not always feasible due to ethical or logistical constraints. Consequently, researchers may consider causal inference and large-scale, observational, retrospective datasets as a bridge between gold standard RCTs and ubiquitous real-world data.[10-13] When observational data are curated to follow a set of assumptions (Figure 1), causal inference can be applied to emulate an RCT, yielding valid treatment effect estimates.[13] Causal ML, in particular, shows promise in leveraging these data since ML approaches excel at extracting signals from multi-modal data such as images, clinical notes, and time-series data. This allows for adjustment on more complex confounders since modalities such as raw images are difficult to incorporate into models using traditional statistical methods due to their lack of well-defined atomic features.

Drawing valid causal inferences requires justifying assumptions in the data and modeling choices. The flexibility of ML increases the risk of erroneous findings. Inappropriately drawing causal conclusions carries risks to medical science and potentially patient safety.[14-16] Here, we build on previous discussions[17-23] and provide practical guidance on using causal ML in observational data. We consider the setting of a binary treatment and outcome to lay the foundation for more complex analyses (eg, multiple treatments).

**IDENTIFICATION CHALLENGES: ARE CAUSAL INTERPRETATIONS POSSIBLE?**

Causal inference aims to estimate the effect of one variable, ie treatment, on another, ie outcome, (eg, off-label use of amiodarone for restoring hemodynamic stability).[24] When estimating causal effects from observational data, treatments are often non-randomly assigned, confounding direct comparisons of outcomes. The causal relationships between the treatment, outcome, and other relevant variables can be modeled using a causal directed acyclic graph (DAG).[10,20] The structure of a causal DAG implies a set of independence conditions between variables that can be used to generate *identifiability assumptions* (Figure 1), or assumptions that determine when a causal interpretation of treatment effects is tenable. Constructing an accurate DAG with input from domain experts is therefore a crucial first step in ensuring meaningful causal conclusions.

ML introduces new challenges in verifying causal assumptions. By default, most ML approaches are optimized for predictive performance. Thus, ML models may leverage spurious associations to make predictions even when an underlying causal relationship exists.[25-27] To illustrate, we review the main identifiability assumptions in causal inference, describe how ML may obscure violations of these assumptions, and offer recommendations for mitigation. These assumptions are defined for some population of interest, from which we assume the data are sampled.

**Overlap.** To satisfy the overlap assumption, every patient should have a non-zero probability of receiving the treatment. Overlap can be assessed, in part, by fitting a *propensity score model*[28] to predict treatment assignment (eg, medication administration, note that we do not predict the outcome here) given variables that could influence treatment assignment according to the causal DAG. If the model is well-specified (ie, captures the underlying relationship between the covariates and treatment assignment), high discriminative performance implies low overlap, and poor discriminative performance indicates more randomness in treatment assignment. If the model is poorly specified, near-random discriminative performance could simply indicate underfitting. Issues with model specification can be mitigated, in part, using techniques from standard supervised ML, such as cross validation for model selection. However, to more thoroughly evaluate overlap, one should consult with domain experts. Modeling treatment assignment is best done in collaboration with those responsible for assigning the treatments; such discussions can help identify settings where overlap is impossible.

**Conditional Exchangeability (No Unobserved Confounding).** The assumption of no unobserved confounding (ie, variables that affect the treatment and outcome) is unverifiable in observational data.[29] Identifying confounders generally requires domain expertise. If potential confounders are unavailable in the data (eg, social determinants of health or patient preferences[30-32]), a causal interpretation may be untenable if the influence of these confounders could flip the conclusions. Even if unmeasured confounders are believed to be small in magnitude, they should be acknowledged as a limitation when communicating the findings. As further assurance, one can employ *sensitivity analyses* that quantify the magnitude of an unobserved confounder necessary to change the conclusions.[33,34]

**SUTVA.** The stable unit treatment value assumption (SUTVA) requires that all patients receive the same "version" of treatment (consistency), and that no patient's outcome is affected by others' treatment assignments (non-interference). Consistency violations may arise due to hidden variation in the treatment variable, eg dose strength. When sources of treatment variation are inadvertently treated as confounders (eg, dose strength), subsequent adjustment can wash out the treatment effect. Interference may occur when treating one patient can affect outcomes for other patients, as seen in the management of nosocomial infections, where decreasing the risk of transmission in one patient may improve outcomes for other patients. Here, hospital room IDs of infected patients may be predictive of infections in subsequent occupants, and an ML model may subsequently use room ID to predict outcomes. In practice, ML practitioners with limited clinical expertise may overlook the causal significance of variables such as dose strength or room ID and include them in model development if they contribute to high predictive performance. However, the utility of dose strength rate would suggest that it should be incorporated into the treatment definition. The utility of room ID would be due to interference, rather than the treatment. Without domain expertise, researchers risk mistaking predictive associations for sources of treatment effect heterogeneity or causal factors.

**Takeaway.** It is essential to first define a causal DAG. One must then examine the data for potential violations of identifiability assumptions using domain expertise and empirical checks. The flexibility of ML models increases the risk of overfitting to spurious signals that may obscure violations of the assumptions. Careful selection of the confounders, treatment/outcome definitions, and associated models in collaboration with domain experts can help mitigate risks of assumption violations or quantify their potential impacts.

## MODELING CHOICES: ESTIMATING TREATMENT EFFECTS

We discuss modeling choices with respect 1) which covariates to include and 2) which modeling approach to use. For ML practitioners, it is tempting to incorporate all observed covariates as predictors into the model to adjust for confounding. However, irrelevant variables may be spuriously correlated with the outcome. Additionally, some variables may be causal mediators, ie, intermediate variables caused by the treatment that influence the outcome. On a causal DAG, these occur along the path from the treatment to the outcome. Adjusting for mediators dilutes estimates of treatment effects, as the treatment may not get “credit” for changes in patient outcomes. One can avoid including mediators by consulting the DAG and ensuring that all predictors are measured before the treatment and follow-up.

We focus our discussion of modeling approaches on metalearners,[35] a popular causal ML approach. These methods allow users to plug in “base” models of arbitrary complexity (eg, logistic regression, random forests, multi-modal foundation models) and estimate treatment effects based on combinations of the outputs of these models. The base model is generally chosen based on factors such as the quantity, dimensionality, and modalities (eg, images) of the data. We discuss two classes of methods for combining base model outputs: (i) indirect and (ii) direct estimators, and their implications on the accuracy of treatment effect estimates (Figure 2).

Indirect estimators estimate the outcome under each treatment assignment separately given the covariates, then take the difference. The S-learner does so with one model, incorporating the treatment assignment as an input covariate.[35] The T-learner fits one model per treatment group.[35] The simplicity of these approaches is appealing since they use supervised ML models to predict outcomes, given covariates and treatment. However, they are fragile. Since the treatment effect is calculated as a difference between two separate predictions, errors in either prediction can be magnified and distort the final estimate. For example, when using high-dimensional data, techniques that simplify the model, such as regularization, can shrink away treatment heterogeneity. This can flatten one or both outcome estimates, which can conceal meaningful treatment heterogeneity or lead to the premature conclusion that the treatment has little/no effect.[35]

Direct estimators use the outputs of base models to model the treatment effect explicitly in a second step. Since only one outcome is ever observed for each individual, the treatment effect cannot be used as a regression target. Direct approaches differ in how they address this issue, ranging from methods that construct pseudo-outcomes (eg, X-learner,[35] DR-learner[36]) to methods that isolate variation unexplained by the treatment assignment or covariates alone (eg, R-learner,[37] double machine learning,[38] orthogonal statistical learning[39]). Although errors in base models can still bias the treatment effect estimate, direct methods are often more robust to these errors than indirect methods.[35,36]

**Takeaway.** When choosing covariates to include, consulting the causal DAG can reduce the risk of including irrelevant variables and mediators. The method of combining base model outputs impacts the quality of the treatment effect estimates. For indirect methods, the accuracy of the final treatment effect estimates depends directly on the correctness of both outcome predictions. Direct methods reframe the problem, using the outputs of base models to estimate the treatment effect directly in a second step. Indirect methods are attractive for their simplicity. However, if the accuracy of base models is a central concern, direct methods may be preferable.

## EVALUATION: HYPOTHESES, NOT FINDINGS

Evaluation in standard supervised ML uses a held-out “test set” of data to compare predicted outcomes to observed outcomes. In causal inference, we only observe the outcome under the assigned treatment. The standard ML paradigm does not apply, as there is no “true” treatment effect for validation.

Investigators can instead check whether the estimates behave in ways consistent with detecting causal signals. One strategy is a global null analysis, which uses a synthetic treatment with no true effect.[12] Because the expected effect is zero, estimated effects can be directly evaluated, similar to supervised ML. To complement a global null analysis, one can perform a negative control analysis (ie, considering a treatment with no effect based on expert knowledge) or plasmode simulation[40] when feasible. Another strategy is to compare estimates across different modeling approaches and assess whether the sign and magnitude of results are stable.[41] Since each modeling approach encodes slightly different assumptions, stability builds confidence in the results. These analyses do not supersede evaluation through RCTs or grounding in real-world causal mechanisms. However, they can provide support for causal hypotheses for future study.

**Takeaway.** Gold-standard validation is generally infeasible in observational causal inference. Causal ML results from observational data should be treated as causal *hypotheses* rather than definitive findings. These hypotheses can then be prioritized for testing in RCTs or evaluated against plausible physiological or biological mechanisms.

## TOWARD RESPONSIBLE USE

Advances in causal ML create opportunities to leverage large-scale multi-modal observational data for clinical research. Without well-justified causal assumptions and ML modeling choices, analyses may yield conclusions that are misleading. We propose a roadmap (Figure 3) that organizes recommendations for the usage of causal ML in observational data into questions addressing causal assumptions, modeling, and evaluation. Although not exhaustive, we offer a foundation for interdisciplinary teams to ground their analyses.

**CONTRIBUTIONSHIP STATEMENT**

JW, SJ, and LAC initially conceptualized the article. The manuscript was written by DT and TC, and edited by DT, TC, SP, RR, AKW, WM, MM, SJ, FDV, LAC, and JW.

**FUNDING STATEMENT**

This work was supported in part by the National Science Foundation (NSF), NSF grant number IIS 2124127. The views and conclusions in this document are those of the authors and should not be interpreted as necessarily representing the official policies, either expressed or implied, of the NSF.

**COMPETING INTERESTS STATEMENT**

The authors have no conflicts of interest to report.

## REFERENCES

1. Obermeyer Z, Emanuel EJ. Predicting the future—big data, machine learning, and clinical medicine. *N Engl J Med*. 2016;375(13):1216-1219. doi:10.1056/NEJMp1606181
2. All of Us Research Program Investigators, Denny JC, Rutter JL, et al. The "All of Us" research program. *N Engl J Med*. 2019;381(7):668-676. doi:10.1056/NEJMsr1809937
3. Johnson AE, Bulgarelli L, Shen L, et al. MIMIC-IV, a freely accessible electronic health record dataset. *Sci Data.* 2023;10(1):1. doi:10.1038/s41597-022-01899-x
4. Veitch DP, Weiner MW, Miller M, et al. The Alzheimer's disease neuroimaging initiative in the era of Alzheimer's disease treatment: A review of ADNI studies from 2021 to 2022. *Alzheimers Dement*. 2024;20(1):652-694. doi:10.1002/alz.13449

5. Hong C, Pencina MJ, Wojdyla DM, et al. Predictive accuracy of stroke risk prediction models across black and white race, sex, and age groups. *JAMA*. 2023;329(4):306-317. doi:10.1001/jama.2022.24683
6. Buell KG, Spicer AB, Casey JD, et al. Individualized treatment effects of oxygen targets in mechanically ventilated critically ill adults. *JAMA*. 2024;331(14):1195-1204. doi:10.1001/jama.2024.2933
7. AlDubayan SH, Conway JR, Camp SY, et al. Detection of pathogenic variants with germline genetic testing using deep learning vs standard methods in patients with prostate cancer and melanoma. *JAMA*. 2020;324(19):1957-1969. doi:10.1001/jama.2020.20457
8. Holste G, Oikonomou EK, Tokodi M, et al. Complete AI-enabled echocardiography interpretation with multitask deep learning. *JAMA*. 2025;334(4):306-318. doi:10.1001/jama.2025.8731
9. Kraemer HC, Kupfer DJ. Size of treatment effects and their importance to clinical research and practice. *Biol Psychiatry*. 2006;59(11):990-996. doi:10.1016/j.biopsych.2005.09.014.
10. Pearl J. *Causality: Second Edition*. Cambridge University Press; 2009.
11. Yao L, Chu Z, Li S, et al. A survey on causal inference. *ACM Trans Knowl Discov Data.* 2021;15(5). doi:10.1145/3444944
12. Xu Y, Ignatiadis N, Sverdrup E, et al. Treatment heterogeneity with survival outcomes. In: Zubizarreta JR, Stuart EA, Small DS, Rosenbaum PR, eds. *Handbook of Matching and Weighting Adjustments for Causal Inference*. 1st ed. Chapman and Hall/CRC; 2023:445-482. doi:10.1201/9781003102670
13. Feuerriegel S, Frauen D, Melnychuk V, et al. Causal machine learning for predicting treatment outcomes. *Nat Med*. 2024;30(4):958-968. doi:10.1038/s41591-024-02902-1

14. Smith GD, Phillips AN. Confounding in epidemiological studies: why "independent" effects may not be all they seem. *BMJ*. 1992;305(6856):757-759. doi:10.1136/bmj.305.6856.757

15. DeStefano F, Shimabukuro TT. The MMR vaccine and autism. *Annu Rev Virol*. 2019;6(1):585-600. doi:10.1146/annurev-virology-092818-015515

16. Rohrer JM, Schmukle SC, McElreath R. The only thing that can stop bad causal inference is good causal inference. *Behav Brain Sci*. 2022;45:e91. doi:10.1017/S0140525X21000789

17. Petersen ML, van der Laan MJ. Causal models and learning from data: integrating causal modeling and statistical estimation. *Epidemiology*. 2014;25(3):418-426. doi:10.1097/EDE.0000000000000078

18. Thomas L, Li F, Pencina M. Using propensity score methods to create target populations in observational clinical research. *JAMA*. 2020;323(5):466-467. doi:10.1001/jama.2019.21558

19. Angus DC, Chang CH. Heterogeneity of treatment effect: estimating how the effects of interventions vary across individuals. *JAMA*. 2021;326(22):2312-2313. doi:10.1001/jama.2021.20552

20. Lipsky AM, Greenland S. Causal directed acyclic graphs. *JAMA*. 2022;327(11):1083-1084. doi:10.1001/jama.2022.1816

21. Holmberg MJ, Andersen LW. Adjustment for baseline characteristics in randomized clinical trials. *JAMA*. 2022;328(21):2155-2156. doi:10.1001/jama.2022.21506

22. Hernán MA, Wang W, Leaf DE. Target trial emulation: a framework for causal inference from observational data. *JAMA*. 2022;328(24):2446-2447. doi:10.1001/jama.2022.21383

23. Dahabreh IJ, Bibbins-Domingo K. Causal inference about the effects of interventions from observational studies in medical journals. *JAMA*. 2024;331(21):1845-1853. doi:10.1001/jama.2024.7741

24. Schallmoser S, Schweisthal J, von Ehr A, et al. Causal machine learning for assessing the effectiveness of off-label use of amiodarone in new-onset atrial fibrillation. Preprint. Posted online June 26, 2025. medRxiv 25330260. doi: https://doi.org/10.1101/2025.06.25.25330260
25. Jabbour S, Fouhey D, Kazerooni E, et al. Deep learning applied to chest X-Rays: exploiting and preventing shortcuts. In: *Proceedings of the 5th Machine Learning for Healthcare Conference*. Proceedings of Machine Learning Research; 2020:750-782.
26. Gichoya JW, Banerjee I, Bhimireddy AR, et al. AI recognition of patient race in medical imaging: a modelling study. *Lancet Digit Health*. 2022;4(6):e406-e414. doi:10.1016/S2589-7500(22)00063-2
27. Banerjee I, Bhattacharjee K, Burns JL, et al. "Shortcuts" causing bias in radiology artificial intelligence: causes, evaluation, and mitigation. *J Am Coll Radiol*. 2023;20(9):842-851. doi:10.1016/j.jacr.2023.06.025
28. Rosenbaum PR, Rubin DB. The central role of the propensity score in observational studies for causal effects. Biometrika. 1983 Apr 1;70(1):41-55.
29. Groenwold RH, Hak E, Hoes AW. Quantitative assessment of unobserved confounding is mandatory in nonrandomized intervention studies. *J Clin Epidemiol*. 2009;62(1):22-28. doi:10.1016/j.jclinepi.2008.02.011
30. Madden JM, Lakoma MD, Rusinak D, et al. Missing clinical and behavioral health data in a large electronic health record (EHR) system. *J Am Med Inform Assoc*. 2016;23(6):1143-1149. doi:10.1093/jamia/ocw021
31. Chen M, Tan X, Padman R. Social determinants of health in electronic health records and their impact on analysis and risk prediction: A systematic review. *J Am Med Inform Assoc*. 2020;27(11):1764-1773. doi:10.1093/jamia/ocaa143

32. Gallagher PJ, Liveright E, Mercier RJ. Patients' perspectives regarding induction of labor in the absence of maternal and fetal indications: are our patients ready for the ARRIVE trial?. *Am J Obstet Gynecol MFM*. 2020;2(2):100086. doi:10.1016/j.ajogmf.2020.100086
33. Manski CF. Nonparametric bounds on treatment effects. *American Economic Review*. 1990;80(2):319-323.
34. Veitch V, Zaveri A. Sense and sensitivity analysis: simple post-hoc analysis of bias due to unobserved confounding. In: Larochelle H, Ranzato M, Hadsell R, Balcan MF, Lin H, eds. *Advances in Neural Information Processing Systems 33*. Curran Associates, Inc.; 2020:10999-11009.
35. Künzel SR, Sekhon JS, Bickel PJ, et al. Metalearners for estimating heterogeneous treatment effects using machine learning. *Proc Natl Acad Sci U S A*. 2019;116(10):4156-4165. doi:10.1073/pnas.1804597116
36. Kennedy EH. Towards optimal doubly robust estimation of heterogeneous causal effects. *Electron J Stat*. 2023;17(2):3008-3049. doi:10.1214/23-EJS2157
37. Nie X, Wager S. Quasi-oracle estimation of heterogeneous treatment effects. *Biometrika*. 2021;108(2):299-319. doi:10.1093/biomet/asaa076
38. Chernozhukov V, Chetverikov D, Demirer M, et al. Double/debiased machine learning for treatment and structural parameters. *Econom J*. 2018;21(1):C1-C68. doi:10.1111/ectj.12097
39. Foster DJ, Syrgkanis V. Orthogonal statistical learning. *Ann Statist*. 2023;51(3):879-908. doi:10.1214/23-AOS2258
40. Franklin JM, Schneeweiss S, Polinski JM, et al. Plasmode simulation for the evaluation of pharmacoepidemiologic methods in complex healthcare databases. Computational statistics & data analysis. 2014 Apr 1;72:219-26.

41. Xu Y, Bechler K, Callahan A, et al. Principled estimation and evaluation of treatment effect heterogeneity: A case study application to dabigatran for patients with atrial fibrillation. *J Biomed Inform*. 2023;143:104420. doi:10.1016/j.jbi.2023.104420

## Figure Legends

**Figure 1:** Identifiability assumptions in causal inference. We illustrate in the causal directed acyclic graph (DAG): (1) overlap, (2) no unobserved confounders, and (3) the stable unit treatment value assumption (SUTVA), which includes consistency (no hidden variation in treatments) and no interference (an individual's outcome depends only on their own treatment assignment, not others' assignments).

**Figure 2:** Summary of causal ML modeling approaches (metalearners) discussed in this article.

**Figure 3:** Roadmap for the responsible use and interpretation of causal inference in observational clinical data. By checking (1) whether the observational data can support a causal comparison, (2) whether limitations of the chosen ML approach may affect the conclusions, and (3) consistency of the estimates with capturing true causal signals, researchers can gain confidence in using causal ML to generate causal hypotheses on observational data.

## Figures

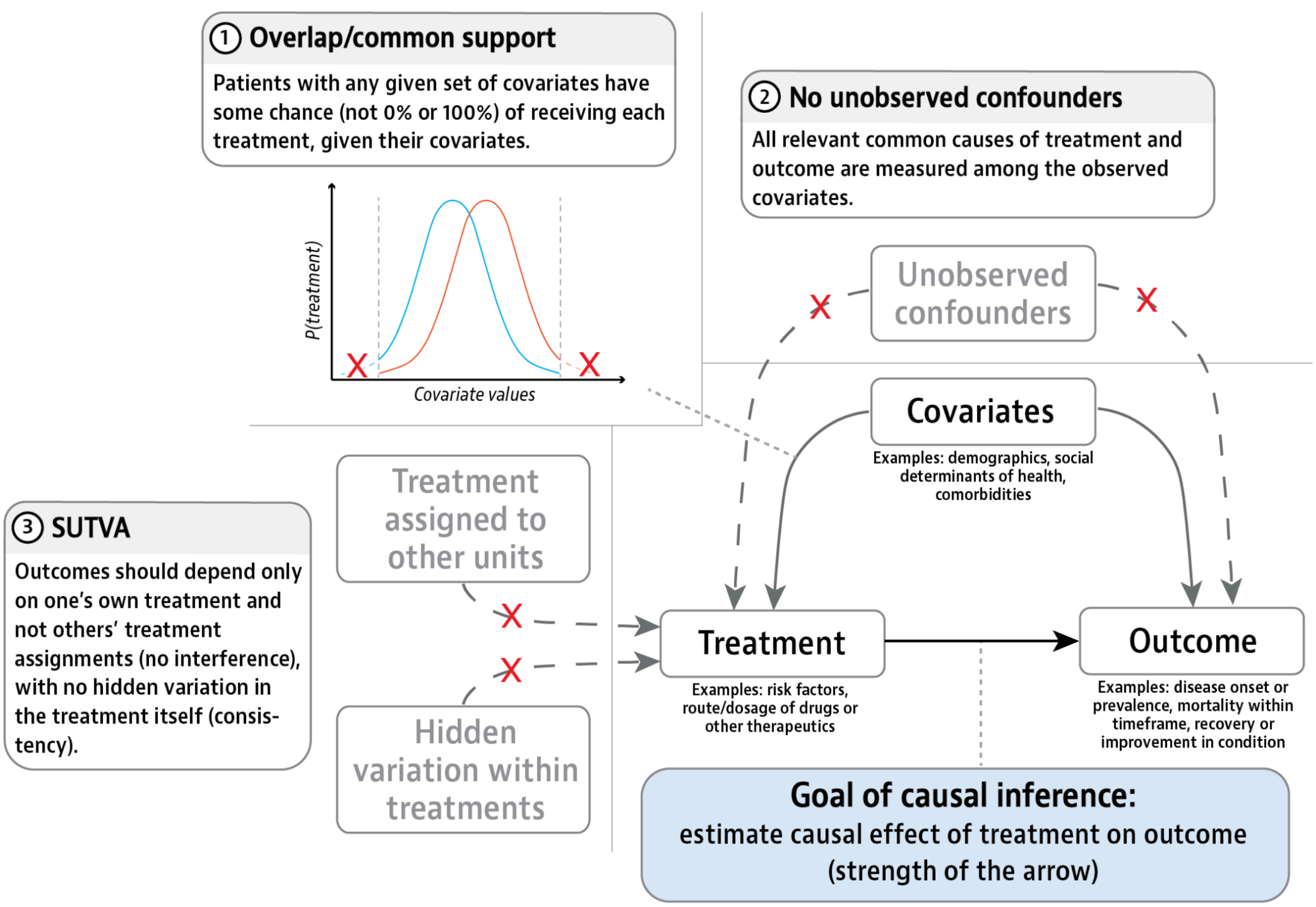

① Overlap/common support
Patients with any given set of covariates have some chance (not 0% or 100%) of receiving each treatment, given their covariates.
P(treatment)
Covariate values
② No unobserved confounders
All relevant common causes of treatment and outcome are measured among the observed covariates.
Unobserved confounders
Covariates
Examples: demographics, social determinants of health, comorbidities
③ SUTVA
Outcomes should depend only on one's own treatment and not others' treatment assignments (no interference), with no hidden variation in the treatment itself (consistency).
Treatment assigned to other units
Hidden variation within treatments
Treatment
Examples: risk factors, route/dosage of drugs or other therapeutics
Outcome
Examples: disease onset or prevalence, mortality within timeframe, recovery or improvement in condition
Goal of causal inference:
estimate causal effect of treatment on outcome (strength of the arrow)

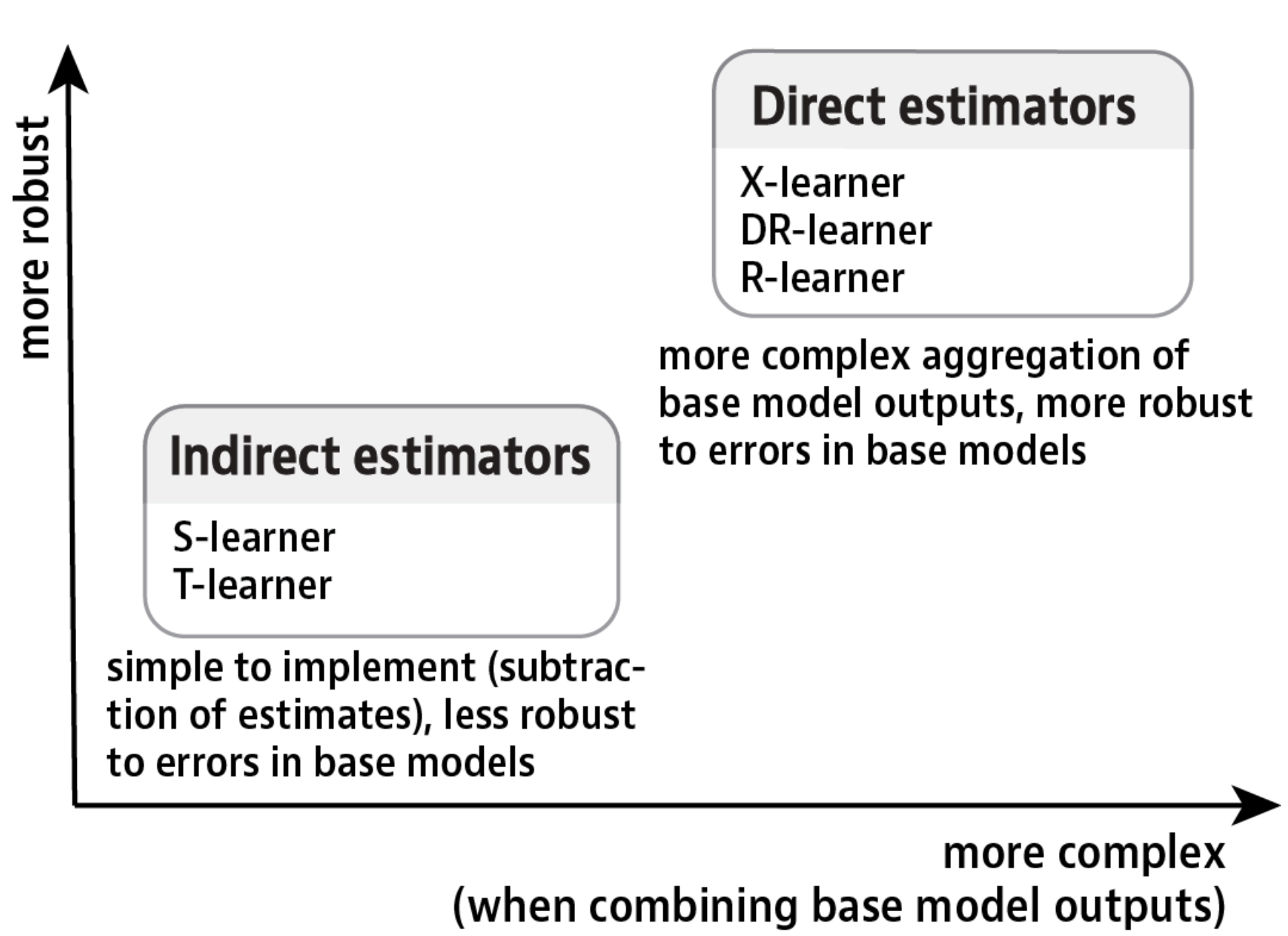

more robust
Direct estimators
X-learner
DR-learner
R-learner
more complex aggregation of base model outputs, more robust to errors in base models
Indirect estimators
S-learner
T-learner
simple to implement (subtraction of estimates), less robust to errors in base models
more complex (when combining base model outputs)

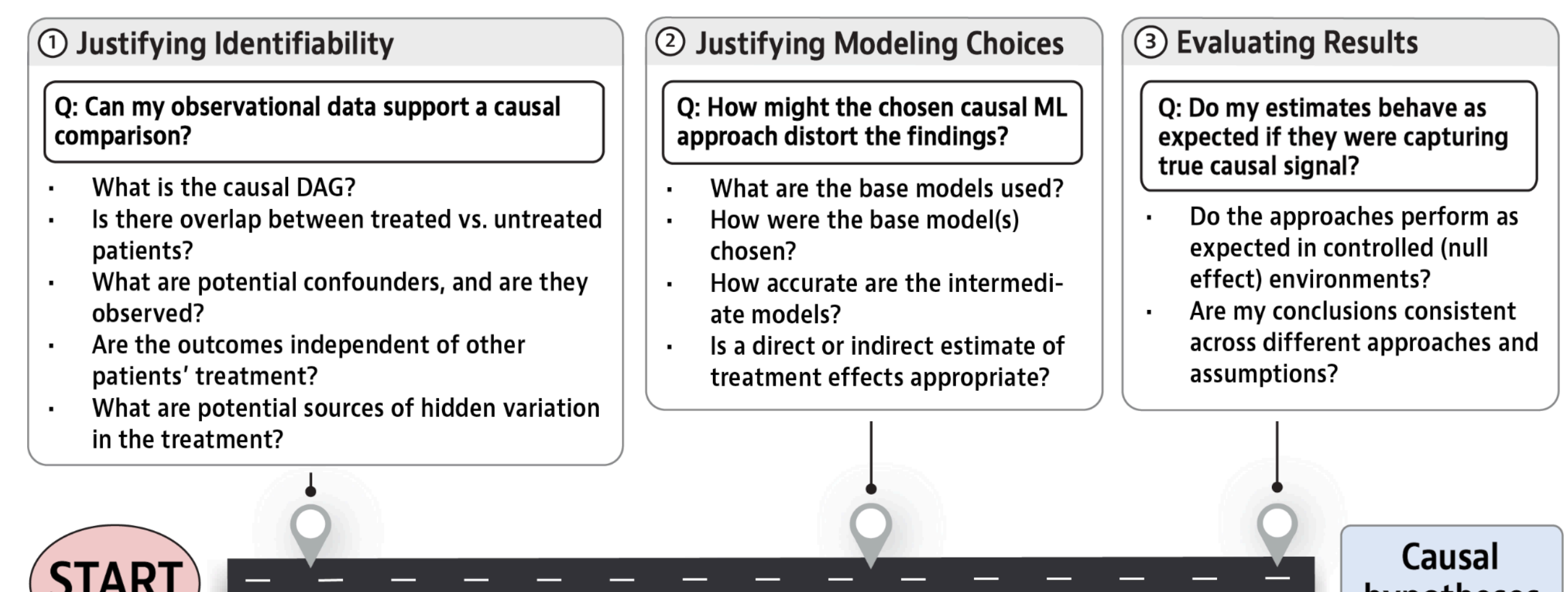

① Justifying Identifiability
Q: Can my observational data support a causal comparison?
What is the causal DAG?
Is there overlap between treated vs. untreated patients?
What are potential confounders, and are they observed?
Are the outcomes independent of other patients' treatment?
What are potential sources of hidden variation in the treatment?
② Justifying Modeling Choices
Q: How might the chosen causal ML approach distort the findings?
What are the base models used?
How were the base model(s) chosen?
How accurate are the intermediate models?
Is a direct or indirect estimate of treatment effects appropriate?
③ Evaluating Results
Q: Do my estimates behave as expected if they were capturing true causal signal?
Do the approaches perform as expected in controlled (null effect) environments?
Are my conclusions consistent across different approaches and assumptions?
START
Causal hypotheses